\documentclass[twoside,11pt]{article}

\usepackage{jmlr2e}
\usepackage{xcolor}
\usepackage{standalone}
\usepackage{algorithm}
\usepackage{algpseudocode}
\usepackage{amsfonts}
\usepackage{amsmath}
\usepackage{comment}
\usepackage{wrapfig}
\usepackage{multicol}
\usepackage{ifthen}
\usepackage{ulem}
\usepackage[sort]{cleveref}
\usepackage{hyperref}

\newcommand{\ceil}[1]{\left\lceil#1\right\rceil}
\newcommand{\floor}[1]{\left\lfloor#1\right\rfloor}

\newlength\defindent
\defindent\parindent
\definecolor{darkgreen}{RGB}{0,150,0}

\newif\ifchange

\long\def\kill#1{\ifchange\raisebox{.9em}{\makebox(0,0)[l]{\color{red}\sout{\detokenize{#1}}}}\fi}
\long\def\replace#1\with#2{\kill{#1}{\ifchange\color{darkgreen}\fi{#2}}}
\long\def\add#1{\replace{}\with{#1}}
\def\undokill#1{{\ifchange\color{red}\fi{#1}}}
\def\undoreplace#1\with#2{\undokill{#1}\ifchange\raisebox{.9em}{\makebox(0,0)[l]{\color{darkgreen}\sout{\detokenize{#2}}}}\fi}
\def\undoadd#1{\undoreplace{}\with{#1}}

\jmlrheading{1}{2020}{1-48}{4/00}{10/00}{\empty}{Vivek Myers and Peyton Greenside}

\ShortHeadings{Hierarchical Batch Bandit Search}{Myers and Greenside}
\firstpageno{1}

\begin{document}
\title{A Hierarchical Approach to Scaling Batch Active Search Over Structured Data}

\author{\name Vivek Myers \email vmyers@stanford.edu \\
       Stanford University\\
       Stanford, CA 94305, USA
       \AND
       \name Peyton Greenside \email pgreenside@alumni.stanford.edu \\
       Stanford University\\
       Stanford, CA 94305, USA}

\editor{What to put here?}


\maketitle


\begin{abstract}
Active search is the process of identifying high-value data points in a large and often high-dimensional parameter space that can be expensive to evaluate. Traditional active search techniques like Bayesian optimization trade off exploration and exploitation over consecutive evaluations, and have historically focused on single or small ($<$5) numbers of examples evaluated per round. As modern data sets grow, so does the need to scale active search to large data sets and batch sizes. 
In this paper, we present a general hierarchical framework based on bandit algorithms to scale active search to large batch sizes by maximizing information derived from the unique structure of each dataset. Our hierarchical framework, Hierarchical Batch Bandit Search (HBBS), strategically distributes batch selection across a learned embedding space by facilitating wide exploration of different structural elements within a dataset. 
We focus our application of HBBS on modern biology, where large batch experimentation is often fundamental to the research process, and demonstrate batch design of biological sequences (protein and DNA). 
We also present a new Gym environment to easily simulate diverse biological sequences and to enable more comprehensive evaluation of active search methods across heterogeneous data sets. The HBBS framework improves upon standard performance, wall-clock, and scalability benchmarks for batch search by using a broad exploration strategy across coarse partitions and fine-grained exploitation within each partition of structured data.
\end{abstract}

\begin{keywords}
Active Search, Bandits, Deep Learning
\end{keywords}

\section{Introduction}

Active search is a canonical problem in machine learning that involves
querying an often unknown or ``black-box" function which assigns labels
to a set of unlabeled data points or features. The goal of active search
is to find examples that maximize this unknown function or that fulfill
some necessary conditions. As modern datasets have grown, active search
methods are increasingly asked to search larger parameter spaces using
batched queries. Our insight is that these large data sets have emergent
structure which is naturally conducive to a hierarchical approach, where
we broadly sample unexplored partitions while
greedily optimizing local search within a partition.
We present a novel hierarchical framework for active search that balances exploration
and exploitation in batch search across structured data. Our approach,
Hierarchical Batch Bandit Search (HBBS), treats the search problem as
one of multi-arm bandit optimization, using a 2-hierarchical approach
to balance exploration across $k$ different partitions. At the top
level, the HBBS framework uses Thompson sampling to explore structurally
different data partitions, and then at the lower level greedily exploits
by optimizing within clusters of structurally similar data
points. We evaluate HBBS by comparing it with standard baseline
methods for active search and bandit
optimization on synthetic and real-world biological datasets, where
there is an acute need to identify sequences of research, diagnostic or
therapeutic interest. We find that HBBS usually outperforms baseline
methods when a dataset has distinct structural
features in its embedding.
\section{Background}

\subsection{Active Search and Bandits}
Active search has been traditionally addressed through techniques like Bayesian Optimization\kill{,} where a surrogate model such as a Gaussian Process (GP) is used in conjunction with an acquisition function such as Upper Confidence Bound (UCB) to intelligently sample new points by balancing exploration and exploitation. GPs are popular but notoriously difficult to scale to large data sets. Proposed solutions include altering the GP model, implementation or acquisition function (\cite{mcintire2016sparse, desautels2014parallelizing, gonzalez2016batch,gardner2018gpytorch}), non-GP Bayesian Optimization techniques (\cite{kathuria2016batched, wang2017batched}), or more recent deep learning-based techniques (\cite{damianou2013deep, snoek2015scalable}). Domain-specific approaches to active search\kill{, particularly in life sciences,} have used generative models to propose diverse inputs to maximize properties of interest (\cite{brookes2018design, gomez2018automatic, killoran2017generating, gupta2019feedback}) or surveyed an ensemble of model\replace{ classes based on}\with{s with} strong domain-specific priors (\cite{biswas2018toward, alley2019unified, yang2019machine}). These methods either are not adaptable or optimized for large batch settings, assume \kill{access to }an oracle, or do not generalize\kill{ to key settings of interest}.

The exploration-exploitation trade-off is also handled in the multi-armed bandit context. \replace{Traditionally, a}\with{A} multi-armed bandit consists of a set of ``arms" $\mathcal{A}$, each with some reward distribution. At each time step, an agent selects an arm $a \in \mathcal{A}$ to pull, observing a single value drawn from its reward distribution, with the end goal of maximizing the total reward of all samples. The standard evaluation metric for bandit algorithms is cumulative regret, and to achieve low regret, a bandit algorithm must balance exploration and exploitation. Thompson sampling is a popular \add{bandit }algorithm \replace{which}\with{that} \replace{has this quality when applied to bandits}\with{achieves this balance} (\cite{thompson1933likelihood}). 

\subsection{Contribution}

Our framework combines the structural embedding of deep learning-based
methods with Thompson sampling to guide a top-level search over
partitions of data, while using \kill{more }exploitative methods to
search within each partition. By viewing the selection between different
clusters of data as a bandit problem, and then using a simpler exploitative
policy within each partition, we can scale to explore
the diverse high-level structural features of complex datasets. In
principle, our hierarchical framework could apply any exploitative
approach within each partition, but for time
and implementation ease we have often found good results with greedy
partition search.

One of the difficulties in assessing active search methods is the
variability in their performance across different tasks and data sets.
This problem is particularly apparent in biology due to the large
diversity of datasets available and experimental objectives that are
pursued. We make available an OpenAI Gym (\cite{brockman2016openai})
environment, ClusterEnv, as a public resource to enable assessing
generalizability beyond the relatively small number of currently
publicly available data sets.

\section{Methods}
\label{sec:meth}

\subsection{Problem statement}
\begin{wrapfigure}[22]{R}{.5\linewidth}
\vspace{-2em}
\scalebox{.88}{
\begin{minipage}{1.08\linewidth}
\begin{algorithm}[H]
    \caption{HBBS[$k, (\mu_0, n_0, \alpha, \beta)$]}
    \label{alg:bucket}
    \begin{algorithmic}[1]
        \Procedure{Act}{$f, \mathcal{D}, M, S$}
        \newsavebox\first\newsavebox\second
        \sbox\first{Observations $\mathcal{D}$, labels $f$}
        \sbox\second{Select $M$ sequences in $S$ to sample}
        \Statex\empty\Comment{\makebox[\wd\second][l]{\usebox\first}}
       	\Statex\empty\Comment{\usebox\second}
        \State{Fit predictor $\hat{f}$ and embedding $\hat{e}$ to $f, \mathcal{D}$}
        \State{k-means cluster $S$ into $S_1\ldots S_k$}
        			\Statex{\qquad\qquad using $\ell_2$ metric induced by $\hat{e}$}
        \For{$i \in \{1\ldots k\}$}
        \def\xi{\mathbf{x}}
        \State{$d_1\ldots d_m \gets S_i\cap\mathcal D$}
        \State{$x_1\ldots x_m \gets f(d_1)\ldots f(d_m)$}
        \State{$\mu \gets \text{mean}~\{x_1\ldots x_m\}$}
        \State{$\tau\gets \Gamma\left(\alpha + {m \over 2}, \beta + \frac{1}{2} \sum_j (x_j - \mu)^2\right.$}\Statex{\hspace{4cm}$\left. + m n_0 \left({(\mu-\mu_0)^ 2 \over 2 (m + n_0)}\right)\right)$}
        \State{$\mathcal{NG}_i\gets \mathcal{N}\left({m  \mu + n_0  \mu_0 \over m + n_0 },\frac{1}{\sqrt{m \tau + n_0 \tau}}\right)$}
        \EndFor
        \State{$A \gets \{\}$}
        \For{$j \in \{1\ldots M\}$}
        	\newbox\tmp
            \State{$\forall i, q_i \sim \mathcal{NG}_i$}
            \State{$b \gets \arg\max_i q_i$} \sbox\tmp{Thompson step}\Comment{\usebox\tmp}
            \State {$a \gets \underset{x \in (S_b \setminus \mathcal{D} ) \setminus A}{\arg\max} \hat{f}(x)$} \Comment{\makebox[\wd\tmp][l]{Greedy step}}
            \State{$A \gets A \cup \{a\}$}
        \EndFor
        \State{\textbf{select $A$}}
        \EndProcedure
    \end{algorithmic}
\end{algorithm}
\end{minipage}
}
\end{wrapfigure}

To formalize the problem\kill{ we are solving}, we model a biological
sequence environment as a triple $E = (S, f, M)$, which includes a collection
of unlabeled sequences $S$, a function $f$ which assigns a label (score) to a sequence, and a batch size $M$. Each $x_i \in S$ is a fixed-length
string and each label $f(x_i) \in [0, 1]$. Agents acting in $E$ propose sets of sequences (batches) to try to optimize the label function over $S$; the value $M$ is the size of these batches. At each time step $t$, an agent acting in $E$ has a set of previously observed sequences $\mathcal{D}_t$, with the ability to see $f(x)$ for any $x \in \mathcal{D}_t$. Based on these observations, the agent selects $M$
elements $z_1 \ldots z_M \in S$ to observe the labels of $f(z_1)
\ldots f(z_M)$. At the next time step, these selections are added to
the set of observations, so $\mathcal{D}_{t+1} = \mathcal{D}_t \cup
\{z_i\}_{i=1}^M$.

\subsection{Metrics}
We use two main metrics to evaluate how well agents find sequences with good labels.

\paragraph{Regret.} First, we use the following cumulative regret metric: For each past selected batch of size $M$, we compute the sum of the differences between
the scores of  the top $\lfloor \rho M \rfloor$ of the selected sequences' scores and those of the top $\lfloor \rho M \rfloor$ of the best possible batch of size $M$ that could be selected. Regret \replace{serves as a proxy for}\with{represents} the distance between each agent and the optimal agent\kill{ at each time step}. Formally, at each time step, we have
regret $r_t = \sum_{n=1}^t R_n^* - R_n$ where $R_n$ and $R_n^*$ are
the sum of the top $\lfloor \rho M \rfloor$ of the labels in a batch drawn from the
remaining sequences at time step $n$, for the agent's selections and
the maximal selections respectively. We focus on the value $\rho = 0.2$ in this paper, \kill{corresponding to }evaluating the top 20\% of each agent's selections.

\paragraph{Time.} One of the most common issues with GPs is their time to fit. As our second metric for evaluating agents, we compare for each method the wall-clock time for a single run at each time step, where a single run is defined as execution of a single hyper-parameter choice\kill{ for that method}.

\subsection{Hierarchical Batch Bandit Search}

We present a 2-level hierarchical model with a bandit component and greedy component. The bandit component employs Thompson sampling with a Gaussian prior.
\Cref{alg:bucket} presents the core HBBS algorithm. Our key insight is
that using an embedding, we can use clustering methods to create the arms of a bandit\replace{. Given this embedding we can determine structurally related clusters containing both observed and unobserved points. }\with{ from structurally related sequences. }At each time step, we refit a predictive model and embedding function to the observed data points $\mathcal{D}_t$. We k-means cluster the observed and unobserved points in the embedding to obtain clusters $S_1\ldots S_k$. We then construct normal-gamma conjugate distributions $\mathcal{NG}_1 \ldots \mathcal{NG}_k$ for the elements of $\mathcal{D}_t$ in each $S_i$. Finally, $M$ times, we sample from each $\mathcal{NG}_i$, denoting the maximum $\mathcal{NG}_b$, and select the unobserved sequence in $S_b$ that maximizes the predictions of our predictive model.

\paragraph{Architecture and Hyperparameters.}
\columnsep .5cm
\begin{wrapfigure}[14]{l}{.4\linewidth}
	\vskip -3.5ex
    \centering
    \includegraphics[scale=0.24]{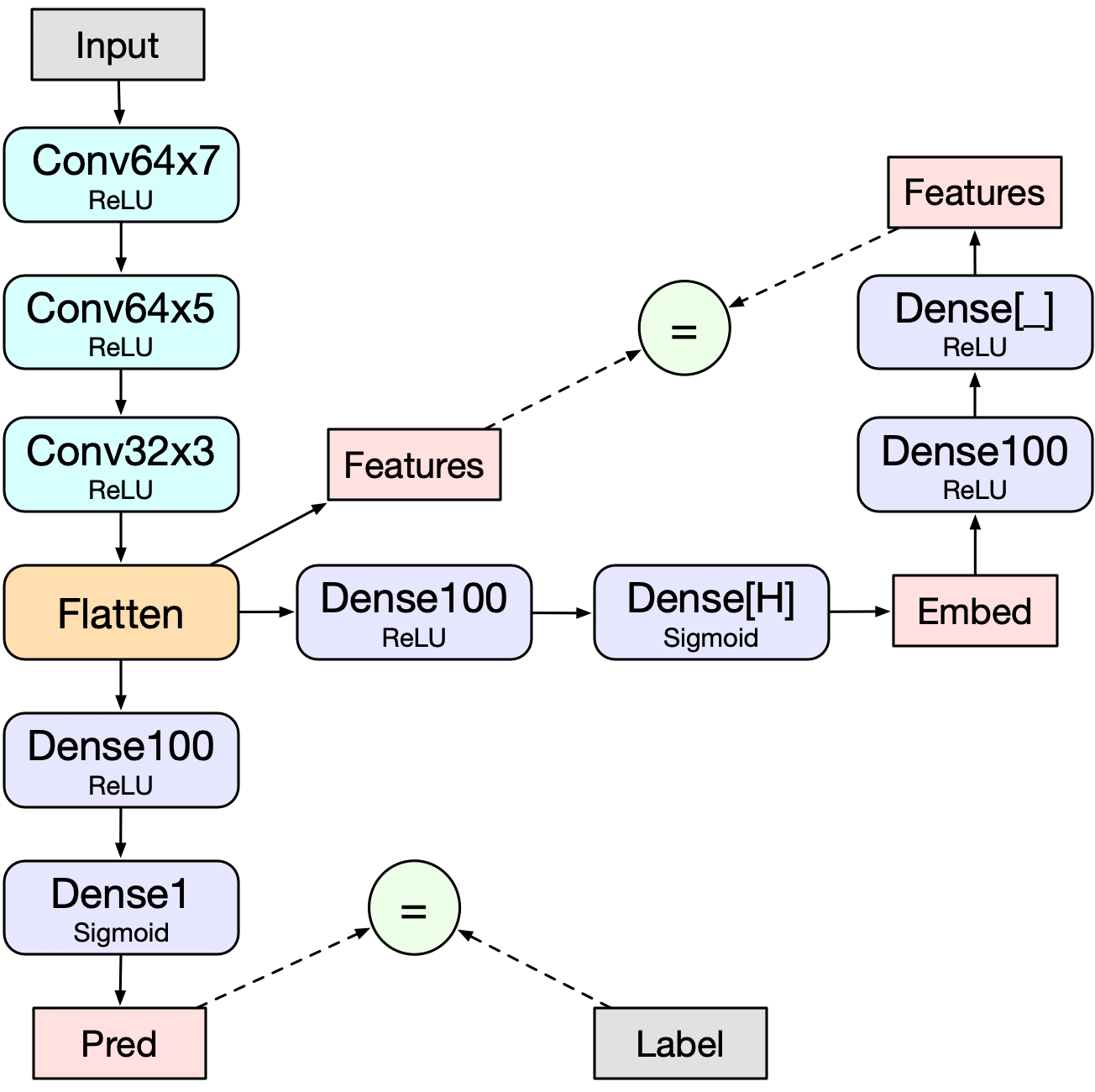}
    \vskip -2ex
    \caption{Model architecture for predictions and dimension $H$ embedding.}
    \label{fig:architecture}
\end{wrapfigure}

We use a deep neural network architecture as a predictive
model and for embeddings (as described in \ref{Architecture}). The Thompson sampling in \replace{the structured bandit approach}\with{HBBS} requires priors $\mu_0, n_0, \alpha, \beta$ for conjugate distributions maintained at each cluster\replace{. The}\with{ and a} parameter $k$ \kill{is also required }\kill{to determine }for the number of clusters \kill{used }at each time step. We found \replace{the HBBS method}\with{HBBS} was \kill{relatively }robust with respect to the prior hyperparameters, and used $\mu_0 = 0.5, n_0 = 10, \alpha = 1, \beta = 1\kill{,}$\kill{ for all runs}.

\subsection{Baselines}
One baseline method is a greedy agent that selects the batch of $M$
sequences maximizing the predictive model
(\Cref{alg:greedy}).  We also compare against a simple
exploration technique that randomly selects $\ceil{\varepsilon M}$ of
the batch of sequences to sample, and samples the remaining $\floor{(1
- \varepsilon)M}$ of the sequences greedily
(\Cref{alg:egreedy}).

\kill{\subsection{GP-UCB}}
We further analyze \kill{the performance of }a batch version of 
\kill{the }GP-UCB \kill{algorithm }(\Cref{alg:gpucb}). We first fit a Gaussian process model to the observed sequences $\mathcal{D}_t$ at each time step, which provides mean and uncertainty predictions $\mu, \sigma$ for each sequence in $S$. Then, we greedily sample \kill{a batch of }$M$ sequences from the upper confidence bound $\mu + \sigma \sqrt{\beta}$ for a \kill{fixed }hyperparameter $\beta$. Every $m$ sequences sampled, we {can }refit the $\sigma$ predictions\kill{ as Gaussian process regression does not require the labels for uncertainty predictions}. The baseline algorithms \replace{can be found}\with{are} in \Cref{sec:algo}.

\subsection{Architecture}
\label{Architecture}
\parbox{\textwidth}{ All agents that require an embedding of prior observed sequences or a predictive model use the same architecture (Figure~\ref{fig:architecture}). For predictions, we fit a model to $\mathcal{D}_t$, the observed sequences with labels, at each time step\replace{. The neural network uses}\with{ using} 1D convolutions\kill{ for these predictions}. Further, many agents require some metric representing the similarity of different sequences\replace{. We construct this metric}\with{, which we construct} by embedding the \replace{layer of the neural network after}\with{output of} the convolutional layers\add{ with an autoencoder}. \replace{Viewing the convolutional layers as extracting the features of sequences corresponding to label}\with{Since the convolutional layers extract features}, the embedding will result in sequences that are similar in label for similar structural reasons clustering closely. The model has several hyperparameters (learning rate, epochs to train, embedding dimension, etc.) detailed in Appendix~\ref{sec:hyper}, which were selected for robust performance across\kill{ several} protein \kill{environments }and \kill{the }artificial cluster environment\add{s}.}

\columnsep .8em
\begin{wrapfigure}[29]{R}{.38\textwidth}
    \centering
    \vskip -4em
    \includegraphics[width=.37\textwidth]{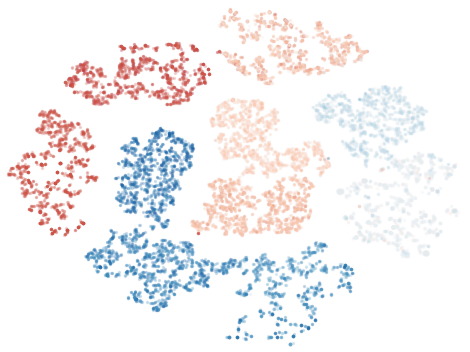}
    \vskip -2.5ex
    \caption{Cluster environment}
    \vskip 1ex
    \label{fig:clust-embed}
    \vskip 0em
    \includegraphics[width=0.37\textwidth]{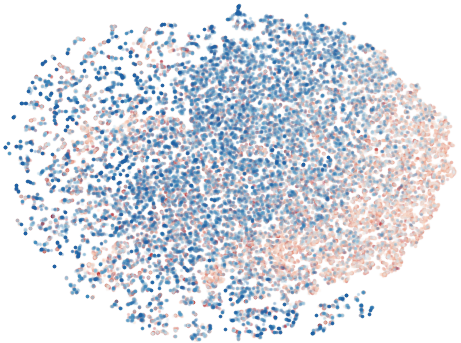}
    \vskip -1.5ex
    \caption{MPRA environment}
    \label{fig:mpra-embed}
    \vskip 1em
    \includegraphics[width=0.37\textwidth]{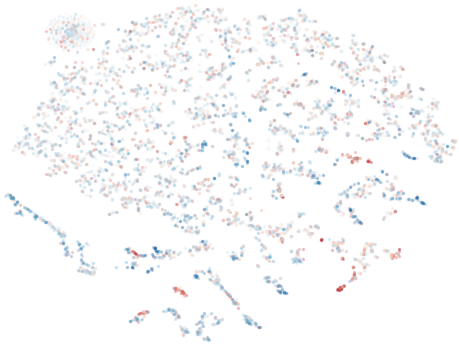}
    \vskip -1.5ex
    \caption{BRCA1 environment}
    \label{fig:brca-embed}
\end{wrapfigure}

\section{Experiments}
\subsection{Environments}
We assessed our approach on one synthetic environment \add{(ClusterEnv) }as well as two real environments: one with DNA sequences\add{ (MPRA)} and one with protein sequences\add{ (BRCA1)}. Our synthetic environment, made publicly available\kill{ as a new Gym environment}, simulates clusters of sequences with related features and labels with \replace{clearly defined}\with{clear} structure. See Appendix~\ref{sec:env} for details\kill{ of our environments}.

\subsection{Setup}
We compared performance of HBBS agents against greedy, $\varepsilon$-greedy, and GP-UCB baselines on the aforementioned environments. We also generated TSNE plots (\cite{maaten2008visualizing}) of the embeddings produced by our architecture when trained on half of the sequences in each environment, plotting the embedding colored by label on the other half. All agents were run for 60 time steps. The artificial environment was run with batch size $M = 100$ while all of the real environments were run with $M = 20$. For computational reasons, the GP-UCB algorithm could not be run on the artificial environment as doing so would require repeated GP evaluations on $\sim$3000 sequences. At the end of each run, we recorded the final Regret(0.2) obtained. For each environment, we randomly selected batches of 32 agents to run concurrently with 32 CPUs and 8 NVIDIA GeForce GTX 1080 Ti GPUs. We repeated this process until each agent had been run for 60 trials or \kill{until }a set amount of time had elapsed. 

Due to some environments taking significantly longer to run than others, we varied this set amount \replace{bempagetween}\with{between} environments so there would be sufficient runs of each agent. We ran agents on the MPRA, BRCA1, and cluster environments with the aforementioned setup for approximately $50, 30, 20,$ and $10$ hours respectively. Final Regret(0.2) values at the last time step for each agent and environment are presented with 90\% confidence intervals. We also compare the wall-clock time of agents on the MPRA environment, averaged across all trials with one representative for each class of agents with the same time complexity.

\columnsep .6em
\begin{wrapfigure}[35]{R}{.46\textwidth}
    \centering
    \vskip -5ex
    \includegraphics[width=1.05\linewidth]{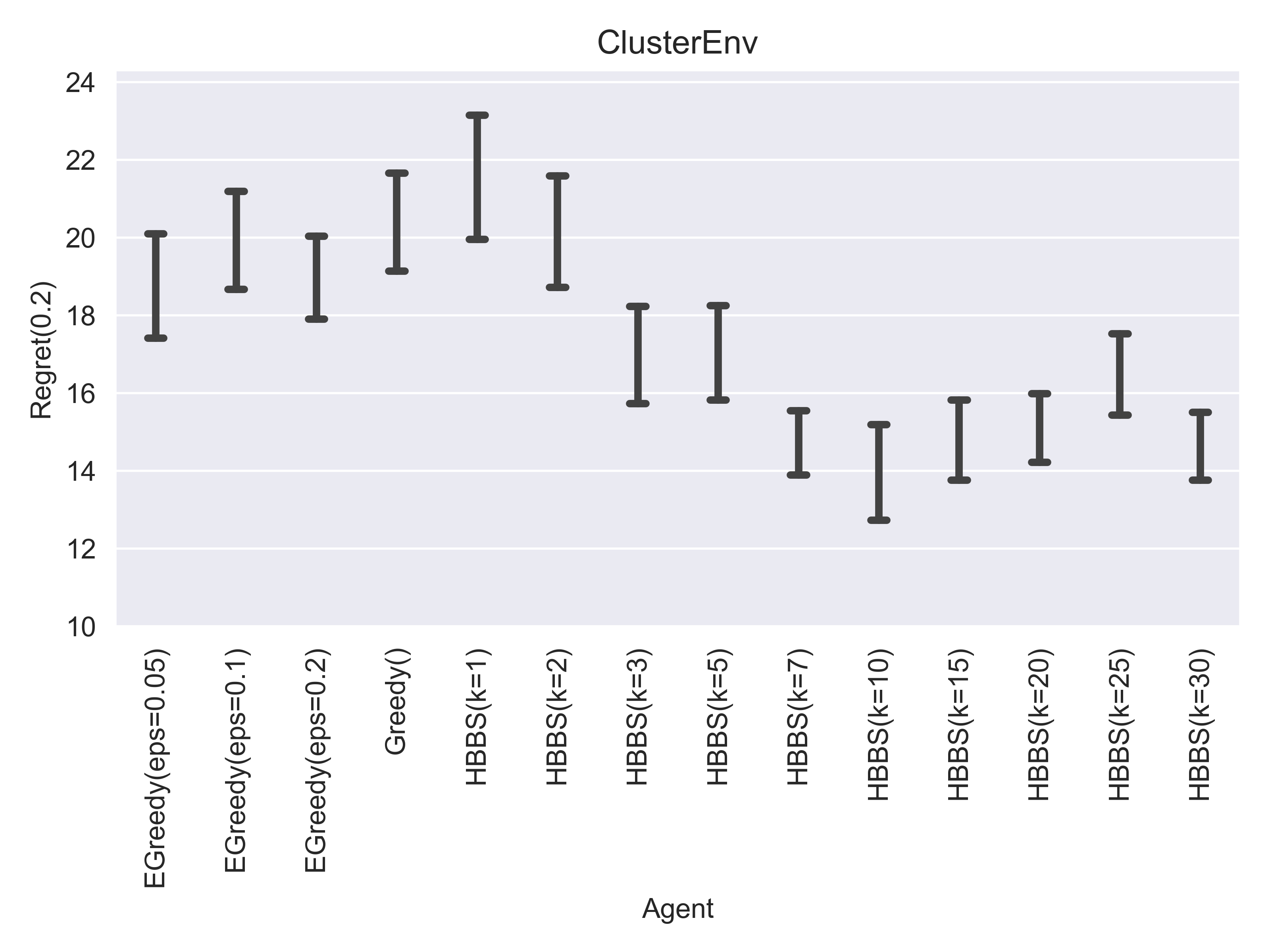}
    \vskip -1.5em
    \caption{Cluster environment}
    \vskip .5ex
    \label{fig:clust-cfinal}
    \includegraphics[width=1.05\linewidth]{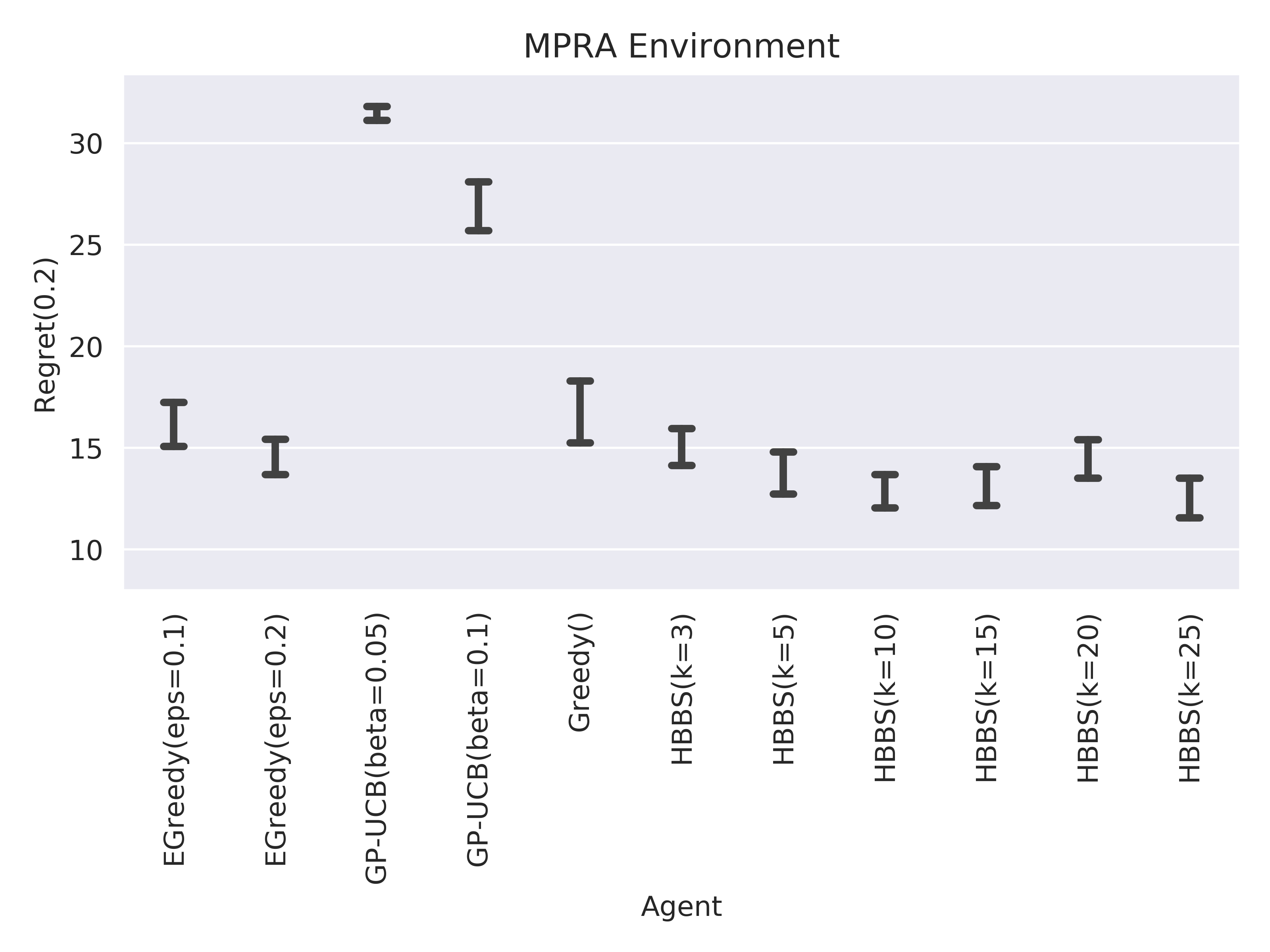}
    \vskip -1.5em
    \caption{MPRA environment}
    \vskip .5ex
    \label{fig:mpra-results}
    	    \includegraphics[width=1.05\linewidth]{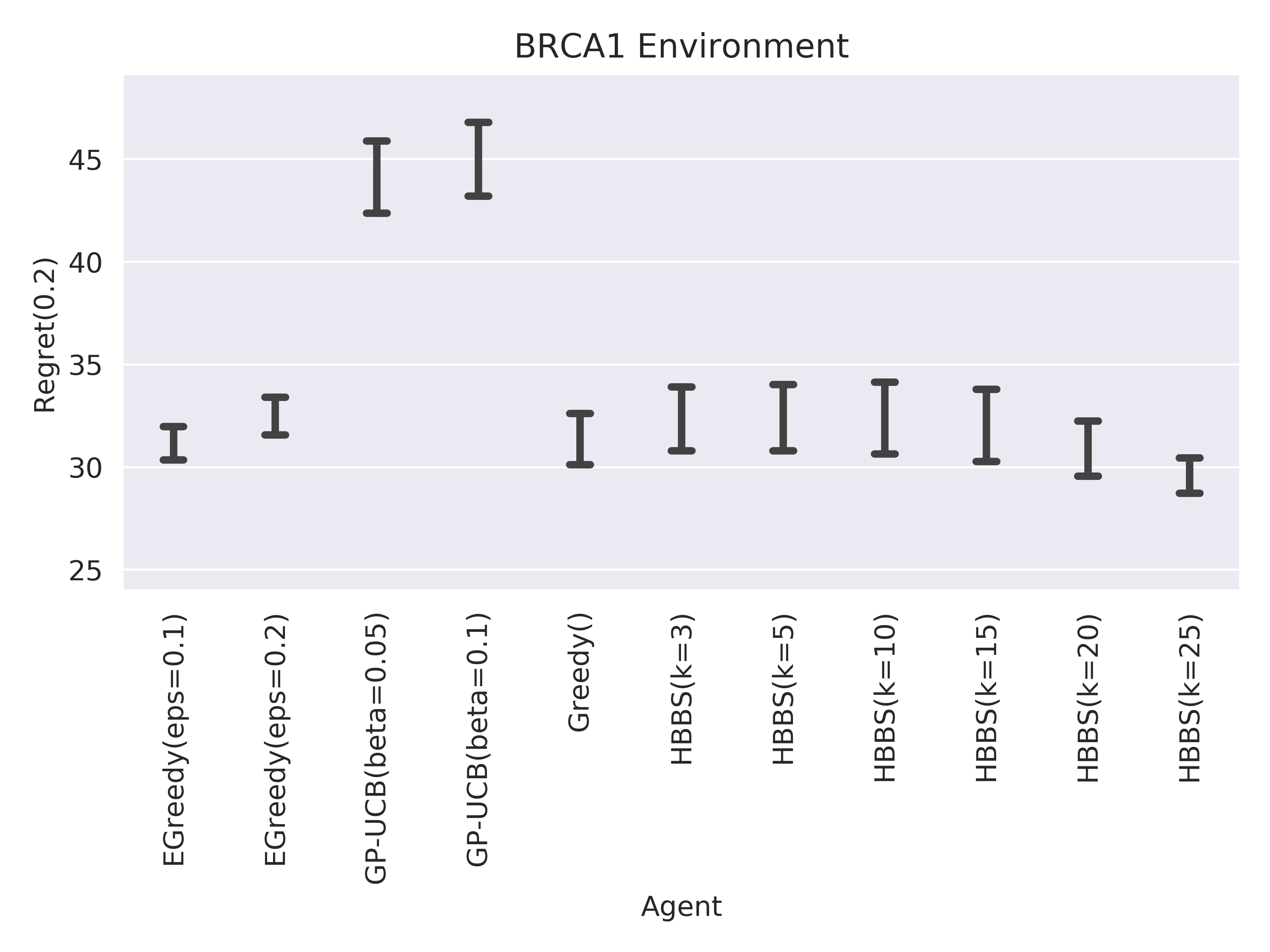}
   	\vskip -1.5em
    \caption{BRCA1 environment}
    \label{fig:brca-results}
\end{wrapfigure}

\subsection{Results}
We present sequence embeddings, colored by label (blue high) in \Cref{fig:mpra-embed,fig:clust-embed,fig:brca-embed}.
We also compare the final regret of HBBS, $\varepsilon$-greedy, and greedy agents in \Cref{fig:mpra-results,fig:clust-cfinal,fig:brca-results}.
See Appendix~\ref{sec:eff} for \kill{the results of }efficiency experiment\replace{s}\with{ results}.

\subsection{Discussion}
\paragraph{ClusterEnv.}
	The clear separation of the embedding into multiple clusters \kill{of similar-scoring sequences }indicates that HBBS may perform well on this environment.
	
	Indeed, HBBS performs optimally near the number of actual clusters in the environment.
\paragraph{MPRA.} 
	In the embedding there are multiple overlapping regions of good and bad sequences\kill{, where good is considered a high expression or reporter level label}.
	
	The HBBS agents with $k$ between $5$ and $15$, as well as $k=25$ perform \kill{the }best\kill{ on this environment}.
\paragraph{BRCA1.}
	The embedding exhibits \kill{very }complex structure, with numerous high scoring clusters.
	
	The HBBS agent with \replace{the highest $k$ value, 25,}\with{$k=25$} performs statistically significantly better than all other agents.

\section{Conclusion}

\leavevmode
\replace{Overall, w}\with{W}e found HBBS to perform the best on all the environments presented with obvious nonlinear structure to their embedding\replace{ and }\with{---}its primary shortcoming \undokill{in performance }is often achieving similar or slightly improved results to $\varepsilon$-greedy. The overall trends we observe in wall-clock time demonstrate HBBS \undokill{can }effectively search\undoadd{es} across tens of thousands of sequences with easy scalability to larger datasets. $\varepsilon$-greedy has been documented as a reliable baseline for similar active search tasks, though its performance is highly dependent on the parameter $\varepsilon$ (\cite{hernandez2017parallel}). Most surprising is the \undokill{relatively }poor performance of GPs, but this can be explained by our observation that methods that \undoreplace{can take advantage of }\with{use }the neural network's predictive capacity consistently outperform those that do not.

In execution time, HBBS can cause methods which are superlinear in time-complexity with respect to the number of sequences in the dataset to run faster. For instance, if HBBS is used with GP-UCB within each partition, by making the partitions arbitrarily small, the $\mathcal{O}(n^3)$ time complexity of GP-UCB can be mitigated. This key insight that HBBS is a general framework that can \replace{give boosts in speed and/or}\with{improve speed or} performance and can be used hierarchically with any method per partition allows its extension to both increasingly large datasets and \undokill{increasingly }specialized search methods.

\newpage
\acks{We would like to acknowledge support for this project from the Stanford CURIS program, the Schmidt Science Fellowship program, and Emma Brunskill and lab members. Thank you to the Google Cloud Platform for supplying research credits for this work.}

\renewcommand{\theHsection}{\Alph{section}}
\appendix

\section{Code}
Our Gym environment and our code for all experiments is publicly available at\\ \mbox{\url{https://github.com/StanfordAI4HI/HBBS}}.

\section{Model Details}
\label{sec:hyper}
We use a learning rate of $\alpha=5\cdot10^{-4}$ with MSE loss at each of the green nodes in Figure~\ref{fig:architecture} for both the autoencoder and prediction portions of the model, and a minibatch size of $100$. We also use an embedding dimension of $H=5$. 

The gray nodes in Figure~\ref{fig:architecture} represent inputs and the red nodes represent output. Input values are one-hot encoded sequences.

\section{Environments}
\label{sec:env}

\begin{wrapfigure}[21]{r}{.53\textwidth}
\scalebox{.88}{
\begin{minipage}{.58\textwidth}
\vskip -4ex
\begin{algorithm}[H]
    \caption{ClusterEnv[$N, n, \sigma, c, \ell$]}
    \label{alg:cluster}
    \begin{algorithmic}[1]
        \Procedure{Generate}{}
        \Statex{\quad} \Comment{Produces sequences $S$ labeled by $f$}
        \State{$S \gets \{\}$}
        \For {$i \in \{1 \ldots N\}$}
        \State{$\mu \sim U(\frac{-1}{2},\frac{1}{2}) $}
        \State{$\xi \sim \sigma U(0,1)$}
        \For {$j \in \{1 \ldots \ell\}$}
        \State{$x_1, x_2, x_3, x_4 \sim \mathcal{N}(0, 1)$}
        \State{$v_j \gets [|x_1|^{1/c}, |x_2|^{1/c}, |x_3|^{1/c}, |x_4|^{1/c}]$}
        \State{$v_j \gets v_j / \sum_i |x_i|^{1/c} $}
        \EndFor
        \State{$u \gets $ distribution over $\{A,C,T,G\}^\ell$}\Statex{\qquad\qquad\qquad with each component one of the $v_j$}
        \For {$j \in \{1 \ldots n\}$}
            \State{$s \sim u$}
            \State{$S \gets S \cup \{s\}$}
            \State{$y \sim \mathcal{N}(\mu, \xi)$}
            \State{$f(s) \gets \frac{1}{1+e^{-y}}$}
        \EndFor
        \EndFor
        \EndProcedure
    \end{algorithmic}
\end{algorithm}
\end{minipage}}
\end{wrapfigure}

\subsection{ClusterEnv Synthetic Environment}
We present a simple artificial environment that mimics the traditional bandit setting with artificial DNA sequences. This environment consists of a set of sequences $\{A, T, C, G\}^\ell$ in $N$ clusters each containing $n$ sequences. Each cluster is generated by first selecting a probability mass function over $\{A, T, C, G\}^\ell$ as well as a Gaussian label distribution, then sampling $n$ sequences (Algorithm~\ref{alg:cluster}). To ensure labels are within $[0,1]$, they are clamped with the sigmoid function. Because labels depend only on the cluster of a sequence, each cluster can be viewed as an arm in a bandit problem, with a reward distribution determined by the labels it contains. Since the distributions are Gaussian, an optimal agent for Gaussian bandits with $N$ arms will also perform well on ClusterEnv if it views the clusters as the $N$ arms. We used the parameters $N=10, n=30000, \sigma=0.1, c=0.2$ for our experiments as this setup yielded a small number of clusters with highly distinct structure and reward distribution.

\subsection{Real Environments}
We also evaluate the agents described in Section~\ref{sec:meth} on real biological sequence environments.

\paragraph{DNA Sequences.}
DNA environments consist of sequences of the form $\{A,C,T,G\}^l$, with an optional additional strand direction identifier, $+$ or $-$. We test on Massively Parallel Reporter Assay (MPRA) sequences of length 150 base pairs (\cite{Urtecho_Insigne}) designed to promote gene expression to higher levels. The MPRA dataset has exponentially distributed scores, so to avoid creating an environment putting undue focus on any particular sequences, we renormalize it to have labels uniformly distributed on $[0,1].$

\paragraph{Protein Sequences.}
Protein sequence environments consist of sequences of 20 different amino acids of the same length representing mutated versions of a protein, along with labels corresponding to the effects of their mutations on their function. We focus on the the protein BRCA1 in this paper by evaluating data from a parallel assay to measure the effects of missense substitutions in the RING domain of BRCA1 on its activity and binding functions (\cite{starita2015massively}).

\section{Efficiency}
\label{sec:eff}
\begin{figure}[H]
	\begingroup
    \centering
    \includegraphics[width=.75\linewidth]{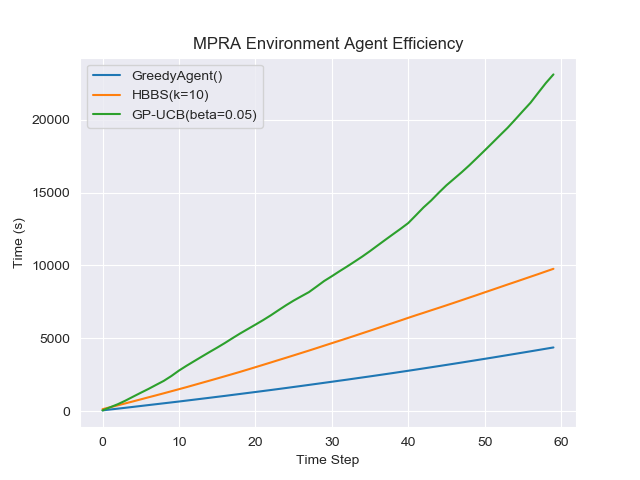}
    \vskip -1ex    
    \caption{Total wall-clock time used at each timestep.}
    \label{fig:time}
    \endgroup
    \vskip 2em
	\paragraph{Discussion.}
	Using HBBS adds overhead which is linear with respect to time step. GP-UCB achieves by far the worst performance, especially at larger time steps. This performance is consistent with the $\mathcal{O}(n^3)$ GP-fitting step which GP-UCB must perform multiple times per action.
\end{figure}

\section{Baseline Algorithms}
\label{sec:algo}

\begin{algorithm}[H]
    \caption{Greedy}
    \label{alg:greedy}
    \begin{algorithmic}[1]
        \Procedure{Act}{$f, \mathcal{D}, M, S$}
        \Statex{\quad} \Comment{Observations $\mathcal{D}$, labels $f$; select $M$ sequences in $S$ to sample.}
        \State{Fit predictor $\hat{f}$ to $f, \mathcal{D}$}
        \State{$A \gets \{\}$}
        \For{$i \in \{1\ldots M\}$}
        \State {$a \gets \underset{x \in (S \setminus (\mathcal{D} \cup A) )}{\arg\max} \hat{f}(x)$}
        \State{$A \gets A \cup \{a\}$}
        \EndFor
        \State{\textbf{select $A$}}
        \EndProcedure
    \end{algorithmic}
\end{algorithm}

\begin{algorithm}[H]
    \caption{$\varepsilon$-Greedy[$\varepsilon$]}
    \label{alg:egreedy}
    \begin{algorithmic}[1] 
        \Procedure{Act}{$f, \mathcal{D}, M, S$}
        \Statex{\quad} \Comment{Observations $\mathcal{D}$, labels $f$; select $M$ sequences in $S$ to sample.}
        \State{Fit predictor $\hat{f}$ to $f, \mathcal{D}$}
        \State{$A \gets$ random sample of $\lceil\varepsilon M\rceil$ sequences in $S - \mathcal{D}$}
        \For{$i \in \{1\ldots \lfloor(1 - \varepsilon) M\rfloor\}$}
        \State {$a \gets \underset{x \in (S \setminus (\mathcal{D} \cup A) )}{\arg\max} \hat{f}(x)$}
        \State{$A \gets A \cup \{a\}$}
        \EndFor
        \State{\textbf{select $A$}}
        \EndProcedure
    \end{algorithmic}
\end{algorithm}

\begin{algorithm}[H]
    \caption{GP-UCB[$\beta, m$]}
    \label{alg:gpucb}
    \begin{algorithmic}[1]
        \Procedure{Act}{$f, \mathcal{D}, M, S$}
        \Statex{\quad} \Comment{Observations $\mathcal{D}$, labels $f$; select $M$ sequences in $S$ to sample.}
        \State{Fit embedding to $f, \mathcal{D}$, use to fit GP $\mu, \sigma$ to $f, \mathcal{D}$}
        \State{$A \gets \{\}$}
        \For{$i \in \{1 \ldots M\}$}
        \State {$a \gets \underset{x \in (S \setminus (\mathcal{D} \cup A) )}{\arg\max} \mu(x) + \sigma(x)\sqrt{\beta}$}
        \State{$A \gets A \cup \{a\}$}
        \If{$m$ divides $i$}
            \State{Refit $\sigma$ with $\mathcal{D} \cup A$}
        \EndIf
        \EndFor
        \State{\textbf{select $A$}}
        \EndProcedure
    \end{algorithmic}
\end{algorithm}

\bibliography{bibliography}

\end{document}